\documentclass[10pt,twocolumn,letterpaper]{article}

\usepackage[pagenumbers]{cvpr} 

\usepackage{makecell}
\usepackage{multirow}
\usepackage{pifont}
\usepackage{soul}
\newcommand{\cmark}{\textcolor{green!60!black}{\ding{51}}}
\newcommand{\xmark}{\textcolor{red!75!black}{\ding{55}}}
\newcommand{\longname}{\textbf{DIS}entangled and \textbf{CO}mposable Skill Mechanisms (DISCO)}
\newcommand{\shortname}{DISCO}

\definecolor{cvprblue}{rgb}{0.21,0.49,0.74}
\usepackage[pagebackref,breaklinks,colorlinks,allcolors=cvprblue]{hyperref}
\usepackage{adjustbox}

\def\confName{CVPR}
\def\confYear{2027}

\title{Uncovering Separable and Composable Skill Mechanisms \\in Vision-Language Models}
\author{
\textbf{Miaoyu Li \qquad Boyang Li} \\
College of Computing and Data Science \\
Nanyang Technological University, Singapore \\
{\tt\small \{miaoyu.li, boyang.li\}@ntu.edu.sg}
}

\begin{document}
\maketitle

\begin{abstract}
Recent work has sought to localize the internal mechanisms underlying visual reasoning skills in vision-language models. These methods largely overlook separability and composability in both mechanism identification and evaluation. We propose separability and composability as two important desiderata for skill localization. Separability requires identified mechanisms to contribute primarily to their target skills while having limited effects on other skills, enabling precise diagnosis and targeted intervention. Composability requires skill mechanisms to be combinable to support complex tasks, allowing a small set of mechanisms to account for broader capabilities. In this work, we propose a multi-skill interpretability framework \longname{}, built around these two properties. For identification, we propose Separable Head Selection, which selects skill-specific attention heads based on their contributions across skills. For evaluation, we are the first to design metrics for measuring separability and composability of identified heads. Extensive experiments on three VLMs and four fundamental visual reasoning skills demonstrate the effectiveness of \shortname{}. It achieves the highest separability, outperforming the strongest baseline by 24.80, 18.38, and 13.75 points across the three models, respectively, while also achieving the highest composability on all three models.
\end{abstract}

\section{Introduction}
\label{sec:introduction}

\begin{table*}[t]
\centering
\caption{
Comparison of head localization methods and skill localization methods. Targeting Skills indicates whether a method targets visual reasoning skills. Separability-Aware Selection describes how each method accounts for separability during component identification.
}
\label{tab:related_work_comparison}

\small
\setlength{\tabcolsep}{3pt}
\renewcommand{\arraystretch}{1.18}

\resizebox{0.8\textwidth}{!}{%
\begin{tabular}{lcccccc}
\toprule

\textbf{Method}
&
\textbf{Granularity}
&
\shortstack[c]{
\textbf{Patching}\\
\textbf{Source}
}
&
\shortstack[c]{
\textbf{Targeting}\\
\textbf{Skills}
}
&
\shortstack[c]{
\textbf{Separability-Aware}\\
\textbf{Selection}
}
&
\shortstack[c]{
\textbf{Composability}\\
\textbf{Evaluation}
}
&
\shortstack[c]{
\textbf{Surface-Form}\\
\textbf{Agnostic}
}
\\
\midrule

NOTICE \cite{golovanevsky2025notice}
&
Head
&
Counterfactual
&
\xmark
&
Implicit
&
\xmark
&
\xmark
\\

V-SEAM \cite{wang2025vseam}
&
Head
&
\mbox{Layer Average}
&
\xmark
&
Implicit
&
\xmark
&
\xmark
\\

VFL \cite{shi2025vfl}
&
Layer
&
Counterfactual
&
\cmark
&
Implicit
&
\xmark
&
\xmark
\\

\midrule

\textbf{Ours}
&
Head
&
Counterfactual
&
\cmark
&
Explicit
&
\cmark
&
\cmark
\\

\bottomrule
\end{tabular}
}

\end{table*}

The identification \citep{tiong2024we,kang2026characterizing}, enhancement \citep{chen2023skill,zhu2025teach}, and mechanistic interpretation \citep{shi2025vfl,schaumloffel2026mechanisms,che2026counting,chen2026skill} of \textit{skills} in Large Language Models (LLMs) and Vision-Language Models (VLMs) have emerged as an important research area. A skill can be defined as a unit of computational capability that serves a single purpose and can be combined to form more complex capabilities. For example, determining the number of apples in an image requires the skill of counting objects. When combined with the positioning skill, which determines the relative positions between objects, they allow the model to answer the question ``how many apples are on the left of the box in the image?" With the understanding of skills, we can efficiently select training data \citep{chen2023skill,zhao2025probing}, steer skill-related internal activations \citep{kang2026characterizing,che2026counting}, and conduct fine-grained evaluation and diagnosis \citep{xu2024multiskill,zhang2025sphere}.

Recent work has sought to localize the network components that implement skills, such as attention heads \citep{jiang2025investigating,schaumloffel2026mechanisms}, layers \citep{shi2025vfl}, and circuits \citep{che2026counting}. We posit that two requirements, separability and composability, are critical to skill localization. \textbf{Separability} requires that the effects of identified components be isolated to the target skill, with minimal impact on other skills; achieving separability would enable precise diagnosis of model errors and targeted skill enhancement with minimal spillover. \textbf{Composability} requires that different skills can be combined to support complex tasks, allowing a small set of skills to parsimoniously account for a broad range of capabilities. 

Existing methods employ a vague notion of separability in component identification, but they neither explicitly formulate this idea nor evaluate separability of identified components. For example, some methods \cite{wang2025vseam,che2026counting,schaumloffel2026mechanisms} select the top-$K$ heads with the largest causal contributions to a skill. This selection excludes lower-ranked heads that still contribute positively to the target skill but may contribute more strongly to other skills. Other methods \cite{golovanevsky2025notice} remove overlapping heads from the candidate sets for different skills. Composability, meanwhile, is entirely overlooked.

To address these limitations, we propose a multi-skill interpretability framework, \longname, which explicitly aims for both properties. We study visual reasoning skills using corresponding VQA tasks. For component identification, we propose Separable Head Selection, which identifies heads that selectively contribute to each skill. For evaluation, we introduce novel metrics to assess the separability and composability of the identified heads.

Separable Head Selection identifies heads for each skill by considering not only their contributions to the target skill but also their contributions to other skills. Specifically, we assign an attention head to Skill A only if its contribution to Skill A exceeds its contributions to any other skill by a separation threshold. Instead of tuning the threshold manually, we find its value by maximizing a separability metric defined on the set of heads selected. This separability is measured as the contribution of these heads to Skill A minus their maximum contribution to any other skill. In plain English, we aim to select a group of heads whose contribution to Skill A exceeds their contribution to any other skill the most, thereby identifying a separable skill mechanism.

Attention heads associated with certain surface forms are often entangled with heads associated with the problem-solving skills, which we aim to identify. For example, if we disrupt a head responsible for outputting the letter A, the network will fail on a multiple-choice question (MCQ) whose correct answer is A. This leads the head localization algorithm to conclude that this surface-form-specific head is important for that question. To solve this issue, we propose Surface-Form-Agnostic Head Scoring, which creates MCQs with multiple option formats (\textit{e.g.}, A/B/C, 1/2/3, and label-free options) as well as images with different styles (natural, cartoon, oil painting). Heads that are highly influential across most surface forms are likely associated with the skill rather than the surface forms. 

Further, we are the first to propose metrics for evaluating the separability and composability of attention heads. The separability metric measures how selectively the identified heads contribute to their target skills, as defined above. The composability metric reflects whether heads for different skills can complement each other to support a task requiring these skills together. Specifically, for a task jointly requiring Skills A and B, we disrupt the heads for Skill A, Skill B, and both skills together, resulting in three performance drops, denoted as $Drop_{A}$, $Drop_{B}$, and $Drop_{A \cup B}$, respectively. Composability is measured by $Drop_{A \cup B} - \mathrm{max}(Drop_{A}, Drop_{B})$. If the identified heads for Skill A and Skill B are indeed complementary, disrupting both sets of heads should incur higher performance drop than disrupting any one set. 


Extensive experiments show that the heads identified by \shortname{} achieve the highest overall separability, outperforming the strongest baseline by 24.80, 18.38, and 13.75 points on Qwen2.5-VL, Qwen3-VL, and Gemma-4, respectively, while also achieving the highest average composability for all three models.

We summarize our main contributions as follows:

\begin{itemize}
\item We propose two critical desiderata for skill localization: separability and composability. \textbf{Separability} requires identified heads to primarily contribute to their target skill while having limited contributions to other skills. \textbf{Composability} requires that identified heads for different skills can be combined to support complex tasks.

\item We propose a multi-skill interpretability framework, \shortname{}, explicitly built around the two desiderata. Furthermore, we design novel metrics to evaluate both the separability and composability of identified heads.

\item We propose two modules for identifying skill-specific heads. \textbf{Separable Head Selection} adopts a separation threshold to select the set of heads whose contribution to the target skill exceeds their contribution to any other skill the most. To mitigate the entanglement between surface-form-specific heads and skill-related heads, we introduce \textbf{Surface-Form-Agnostic Head Scoring}, which computes head contributions using multiple surface forms.
\end{itemize}

\section{Related Work}
\label{sec:related_work}

\label{sec:framework}

\begin{figure*}[t]
    \centering
    \includegraphics[width=\linewidth]{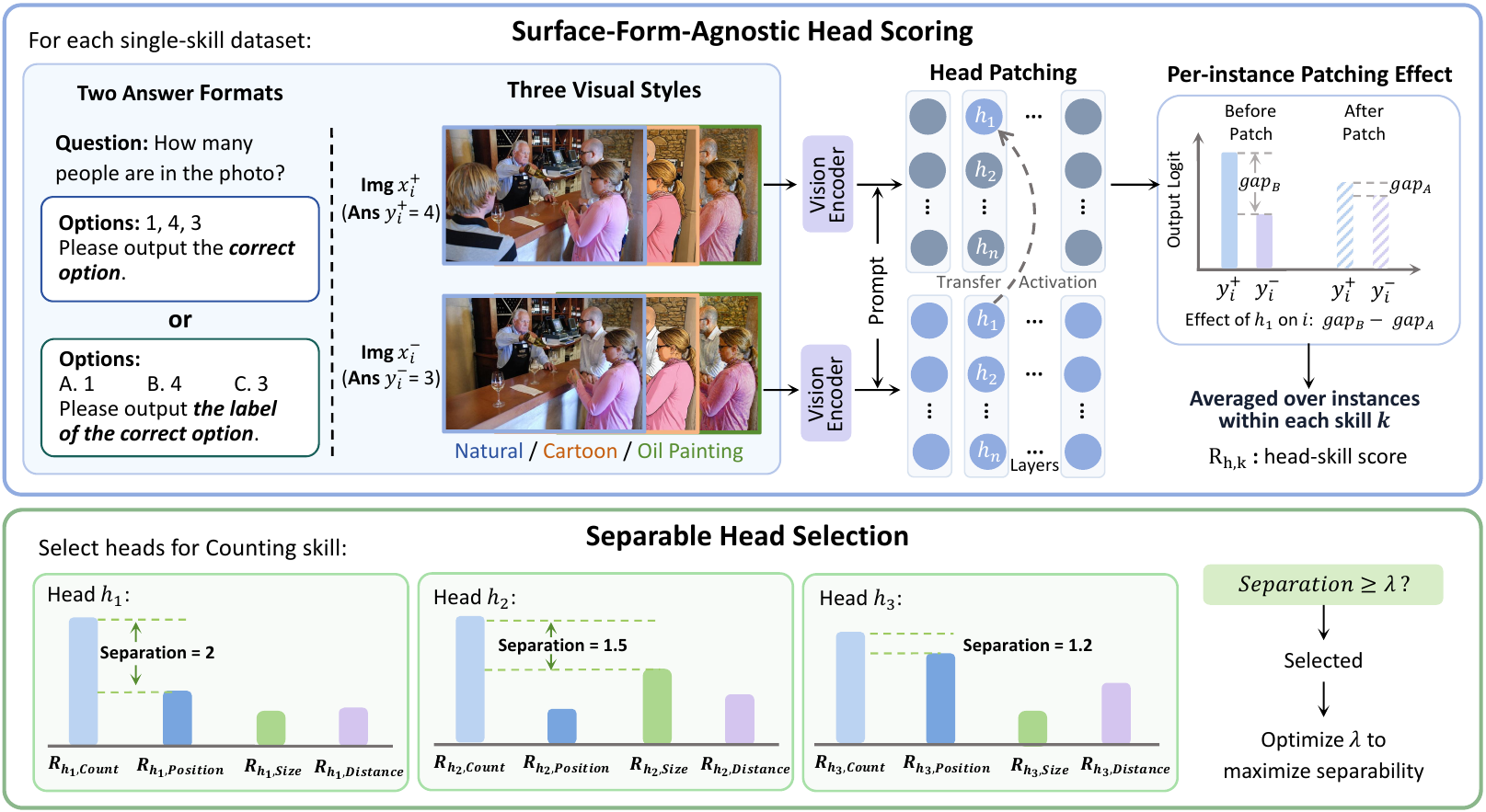}
    \caption{Overview of our head identification pipeline. Surface-Form-Agnostic Head Scoring measures the contribution score of each head to each skill across multiple surface forms. Separable Head Selection optimizes a separation threshold to select skill-specific heads based on their contributions across skills.
}
    \label{fig:framework}
\end{figure*}

\subsection{Skills in VLMs}
\label{sec:related_skills}

Existing studies have investigated skills in VLMs through skill-oriented benchmarking~\citep{johnson2017clevr,hudson2019gqa,zhao2022vl,yuksekgonul2022and,liu2023visual,li2026two,li2026unveiling}, identifying the skills involved in different tasks~\citep{tiong2024we}, analyzing the composition of different skills~\citep{comosition2,composition3}, and localizing skill-related mechanisms~\citep{shi2025vfl,che2026counting,schaumloffel2026mechanisms}.
Current skill localization methods typically identify components using skill data with a single surface form (\textit{e.g.}, a fixed answer format).
For example, VFL~\citep{shi2025vfl} identifies counting and recognition layers using datasets with numerical and Yes/No answers, respectively.
Similarly, Counting Circuits~\citep{che2026counting} identifies counting heads using data with numerical answers, while Object Localization Mechanisms~\citep{schaumloffel2026mechanisms} utilizes data with bounding-box coordinates as the answers.
However, these methods cannot distinguish heads associated with surface forms from those associated with skills~\citep{ismithdeen2025promptception,atabuzzaman2025benchmarking,rosenthal2025unexplored}.
As a mitigation, we identify skill-specific heads using data spanning multiple surface forms.

\subsection{Mechanistic Interpretation in VLMs}
\label{sec:related_causal_localization}
Mechanistic interpretation of VLMs seeks to identify network components supporting specific functions, such as counting~\citep{che2026counting} and modality processing~\citep{golovanevsky2025notice}.
Activation patching~\citep{vig2020causal,meng2022locating,wang2022interpretability,conmy2023towards,zhang2024towards,heimersheim2024use} is widely used to measure the causal contribution of a component by replacing its activation and observing the resulting change in the model output.
To identify modality-related heads, NOTICE~\citep{golovanevsky2025notice} replaces the activation of each head with its counterpart from a counterfactual input.
Similar methods have been applied to identify heads associated with object localization~\citep{schaumloffel2026mechanisms} and counting~\citep{che2026counting}.
VFL~\citep{shi2025vfl} applies this method at the layer level to localize various visual reasoning skills.
V-SEAM~\citep{wang2025vseam} replaces the activation of each head with the average head activation of that layer.

Separability has been an implicit, rather than explicit, consideration in existing methods. V-SEAM~\citep{wang2025vseam} selects the Top-K heads with the largest contributions to each skill, excluding lower-ranked heads that still contribute positively. This exclusion may be deemed to reflect separability, as these lower-ranked heads may contribute more strongly to other skills.
Similarly, VFL~\citep{shi2025vfl} localizes visual reasoning skills by identifying layers with peak contributions.
NOTICE~\citep{golovanevsky2025notice} first identifies candidate heads with high causal contributions to each modality and then removes heads that overlap across different modalities, implicitly favoring separability. 

However, these methods do not explicitly formulate separability as an objective or evaluation metric. Furthermore, they overlook the composability of network components. In contrast, we explicitly consider both separability and composability in head localization and propose evaluation metrics. We highlight the mechanism differences between \shortname{} and other head localization methods and skill localization methods for VLMs in Tab.~\ref{tab:related_work_comparison}.


\section{Methodology}

The proposed framework, \shortname, first identifies skill-specific heads and then evaluates their separability and composability. As illustrated in Fig.~\ref{fig:framework}, the identification stage consists of two modules: (1) \textbf{Surface-Form-Agnostic Head Scoring} quantifies the contribution of each head to each skill across multiple surface forms, and (2) \textbf{Separable Head Selection} utilizes a separation threshold to select heads for each skill based on their contributions across skills. We then design separability and composability metrics for evaluating identified heads. The algorithm of head identification is summarized in the supplementary material.

\subsection{Data Requirement}
\label{sec:notation}

We study a set of visual reasoning skills $K=\{k_1,k_2,k_3,\ldots\}$ using VQA datasets. \shortname{} works on a dataset that labels the data points with the skills required to solve them. Each data point can be denoted as the tuple $(q_i, x_i^+, y_i^+, \mathcal{K}_i)$, where $q_i$ is the question, $x_i^+$ the image, and $y_i^+$ the correct answer. $\mathcal{K}_i \subseteq K$ is a set of skills, such as positioning or counting, required to answer the question. Single-skill data instances can be solved using a single skill (\textit{i.e.}, $|\mathcal{K}_i|$ = 1), whereas multiple-skill data instances require the joint application of more than one skill.

\subsection{Surface-Form-Agnostic Head Scoring}
\label{sec:surface_form_agnostic_scoring}
Heads associated with surface forms are often entangled with heads associated with skills, which we aim to identify. For example, for a question whose correct answer is A, disrupting a head responsible for outputting letter A can also cause the model to answer the question incorrectly. As a result, this head will receive a high contribution score. To mitigate this issue, we propose Surface-Form-Agnostic Head Scoring, which estimates head-to-skill contributions using data with multiple surface forms. Heads that make strong contributions across different surface forms are likely to be associated with the underlying skill.



\vspace{0.1in}
\noindent \textbf{Counterfactual and surface-form data construction.} 
We utilize single-skill data with counterfactual images and multiple surface forms to estimate head-to-skill contributions. For each single-skill instance $(q_i, x_i^+, y_i^+)$, we first construct a counterfactual image $x_i^-$ that yields a different answer $y_i^-$ to the same question $q_i$. We then generate multiple variants of each instance $(q_i, x_i^{\pm}, y_i^{\pm})$ using different surface forms. As shown in Fig. \ref{fig:framework}, each surface form consists of an answer format and a visual style. For answer formats, we require the model to output either the Content Options (\textit{e.g.}, 4 people) or Label Options (\textit{e.g.}, B). For visual styles, we transform the original natural images into cartoon and oil-painting styles while preserving their semantic content. We show the details of data construction in the supplementary material.

\vspace{0.1in}
\noindent \textbf{Position-aware counterfactual activation patching.}
To obtain the contribution of head $h$ to skill $k$, we measure the intervention effect of $h$ through counterfactual activation patching. For the $i$-th data point that requires skill $k$, we first run the VLM on $(q_i,x_i^+)$ and $(q_i,x_i^-)$ separately. Then, we replace the output of $h$ of the $x_i^+$ run using its output from the $x_i^-$ run and compute the rest of the feedforward process as usual. We focus on the output logit margin, $\mathrm{logit}(y_i^-) - \mathrm{logit}(y_i^+)$, of the $x_i^+$ run. The effect of head $h$ on the $i$-th data point is defined as the margin after activation patching of $h$ minus the margin before patching, denoted as $\Delta_{i,h}$. The whole process is shown in Fig. \ref{fig:framework}. A positive $\Delta_{i,h}$ indicates that patching head $h$ shifts the prediction from the original answer $y_i^+$ toward the counterfactual answer $y_i^-$. We treat it as evidence that head $h$ contributes to skill $k$ on the $i$-th data point. A larger $\Delta_{i,h}$ indicates a stronger contribution.

We divide the VLM processing into two stages: context processing, which encodes the network input (the question and the image) before the last token, and answer prediction, which predicts the final answer at the last token position. As the two stages play different roles, it is possible that attention heads relevant to the target skill are different.   At two sets of token positions, including the preceding positions $\mathcal{T}_{\mathrm{preceding}}=\{1,\ldots,T-1\}$ and the final position $\mathcal{T}_{\mathrm{final}}=\{T\}$, we identify two separate groups of attention heads for each skill using the activation patching method described earlier.

\vspace{0.1in}
\noindent \textbf{Surface-form-agnostic contribution score.}
We compute the contribution score of head $h$ to skill $k$ by multiplying its average $\Delta_{i,h}$ across all instances by the proportion of instances with positive $\Delta_{i,h}$. Since the instances span multiple surface forms, a head that contributes to only one surface form will have a low proportion of positive $\Delta_{i,h}$. As a result, it will receive a lower contribution score. 
Furthermore, to enable cross-skill comparison, we standardize the contribution scores across all heads for each skill to have zero mean and unit standard deviation, yielding the final scores $R_{h,k}$ for subsequent head selection.

\subsection{Separable Head Selection}
\label{sec:head_selection}
To identify heads that contribute primarily to the target skill while having limited effects on other skills, we propose Separable Head Selection. Unlike existing methods that implicitly consider separability~\citep{golovanevsky2025notice,wang2025vseam,shi2025vfl}, we explicitly incorporate separability into head selection by comparing head contributions across skills. We first introduce a separation threshold.  A head will be selected as a candidate if its contribution to the target skill exceeds its contribution to other skills by the threshold. Then, we optimize this threshold by maximizing a separability metric defined on these candidate heads. In this way, we can obtain a set of heads with the largest separability. The process is shown in Fig. \ref{fig:framework}.

\vspace{0.1in}
\noindent \textbf{Candidate heads.} We select a set of candidate heads for skill $k$ according to a separation threshold $\lambda$. A head must satisfy two requirements to be selected. First, its target-skill score $R_{h,k}$ must exceed a fixed threshold $z_{\mathrm{th}}>0$. Second, its target-skill score $R_{h,k}$ must be at least $\lambda$ times its largest non-target-skill score $\max_{j\neq k}R_{h,j}$. The resulting candidate head set is:

\begin{equation}
H_k(\lambda) =
\left\{
h \;\middle|\;
R_{h,k} > z_{\mathrm{th}}
\;\text{and}\;
R_{h,k} \geq \lambda \max_{j \neq k} R_{h,j}
\right\}.
\label{eq:candidate_skill_head_set}
\end{equation}
Separation threshold $\lambda$ controls the precision-recall trade-off in head selection, with larger values yielding fewer but more selective heads. Since we analyze head contributions at preceding and final token positions separately, $H_k(\lambda)$ consists of two groups of heads. To obtain the appropriate threshold, we consider $\Lambda=\{\lambda_1,\ldots,\lambda_L\}$ and optimize $\lambda$ through grid search in the next step.

\vspace{0.1in}
\noindent \textbf{Skill-adaptive threshold optimization.}
Different skills may vary in their affinity with other skills, thereby favoring different separation thresholds. In light of this, we optimize the separation threshold separately for each skill. For skill $k$, we select its optimal separation threshold $\lambda_k^*$ by maximizing the separability of $H_k(\lambda)$. 

To compute the separability, we first measure the intervention effects of $H_k(\lambda)$ on different single-skill tasks. We jointly intervene on the two groups of heads in $H_k(\lambda)$ at the preceding and final token positions, respectively. We adopt two types of interventions: (1) Patching, which replaces the outputs of heads in $H_k(\lambda)$ with their corresponding outputs from a counterfactual image input, and (2) Erasure, which zeroes out or reverses the outputs of these heads. The Patching effect on  skill-$j$ task is denoted as $\operatorname{Patch}(H_k(\lambda),j)$ and measured as the fraction of instances whose predictions change to the counterfactual answers after patching. The Erasure effect is denoted as $\operatorname{Eras}(H_k(\lambda),j)$ and measured as the fraction of instances whose predictions change from correct to incorrect after erasure. However, intervening on randomly selected heads can also produce non-negligible effects. To obtain the effect truly resulted from the selected heads $H_k(\lambda)$, we subtract a matched random baseline from $\operatorname{Patch}(H_k(\lambda),j)$ and $\operatorname{Eras}(H_k(\lambda),j)$. For this baseline, we randomly select heads matched to $H_k(\lambda)$ in number, layer range, and intervention token positions. We denote the resulting adjusted effects as $\operatorname{Adjust}_{\mathrm{Patch}}(H_k(\lambda),j)$ and $\operatorname{Adjust}_{\mathrm{Eras}}(H_k(\lambda),j)$.

Then, separability is defined as the effect on the target skill $k$ minus the largest effect on any other skill $j\neq k$. For $\mathrm{a}\in\{\operatorname{Patch},\operatorname{Eras}\}$, the separability is: 

\begin{equation}
\begin{aligned}
\operatorname{Sep}_{\mathrm{a}}\left(H_k(\lambda)\right)
&= \operatorname{Adjust}_{\mathrm{a}}\left(H_k(\lambda),k\right) \\
&\quad - \max_{j\neq k}\operatorname{Adjust}_{\mathrm{a}}\left(H_k(\lambda),j\right).
\end{aligned}
\label{eq:separability}
\end{equation}
A high separability score indicates that the identified heads have a strong effect on the target skill beyond random intervention while having limited effects on the other skills.

Finally, we select the optimal threshold via grid search by maximizing the average of the two separability scores:
\begin{equation}
\begin{aligned}
\lambda_k^{*}
&= \arg\max_{\lambda\in\Lambda} \frac{1}{2} \big(\mathrm{Sep}_{\mathrm{Patch}}(H_k(\lambda)) \\
&\qquad\qquad\quad + \mathrm{Sep}_{\mathrm{Eras}}(H_k(\lambda))\big).
\end{aligned}
\label{eq:optimal_separation_strictness}
\end{equation}
The resulting $H_k(\lambda_k^{*})$ constitutes the final heads for skill $k$. 

\subsection{Evaluation Metrics of Separability and Composability}
Beyond identifying skill-specific heads, we design novel metrics to evaluate the two desiderata of separability and composability. Separability metric assesses whether the identified heads contribute strongly to the target skill but weakly to other skills, while composability metric assesses whether heads of individual skills can complement each other to support corresponding multi-skill tasks.

\vspace{0.1in}
\noindent \textbf{Separability.}
Beyond the single-skill separability metric defined in Eq.~(\ref{eq:separability}), we also evaluate separability on out-of-distribution (OOD) multi-skill tasks. Since their data do not contain counterfactual image pairs, we only use Erasure intervention. For a task jointly requiring skills $A$ and $B$, we define the relevant head combination as $H_{\mathrm{rel}}=H_A\cup H_B$ and the irrelevant combination as $H_{\mathrm{irr}}=H_C\cup H_D$, where $C$ and $D$ are two skills not required. However, the two combinations may contain different numbers of heads, and intervening on more heads naturally tends to produce stronger effects. In light of this, we also use the adjusted Erasure effects, computed by subtracting a matched random baseline to mitigate the influence of head count. For a fixed task $A+B$, we abbreviate $\operatorname{Adjust}_{\mathrm{Eras}}(H,A+B)$ as $\operatorname{Adjust}_{\mathrm{Eras}}(H)$. We define the multi-skill separability as
\begin{equation}
\operatorname{Sep}(A+B)
=
\operatorname{Adjust}_{\mathrm{Eras}}(H_{\mathrm{rel}})
-
\operatorname{Adjust}_{\mathrm{Eras}}(H_{\mathrm{irr}}).
\label{eq:multiskill_separability}
\end{equation}
A high score indicates that the identified heads have strong effects when their associated skills are required by the task but weak effects when they are irrelevant.

\vspace{0.1in}
\noindent \textbf{Composability.}
We measure head composability in terms of complementarity. If the heads identified for two skills complement each other in supporting a multi-skill task, jointly intervening on these heads should produce a larger effect than intervening on the heads of either skill alone. Since joint intervention naturally involves more heads, we use randomly selected heads to control for differences in head count. Specifically, we define composability as:
\begin{equation}
\begin{aligned}
\operatorname{Comp}(A+B)
&= \operatorname{Eras}(H_{\mathrm{rel}}) \\
&\quad - \max\left\{
\begin{gathered}
\operatorname{Eras}(H_A \cup R_B),\\
\operatorname{Eras}(H_B \cup R_A)
\end{gathered}
\right\}.
\end{aligned}
\label{eq:composability_rel}
\end{equation}
where $R_A$ and $R_B$ denote randomly selected heads according to $H_A$ and $H_B$, respectively. $|H_{rel}|=|H_A \cup R_B|=|H_B \cup R_A|$. A positive composability indicates that the heads of the two required skills can complement each other to support the multi-skill task. 

\section{Experiments}
\label{sec:experiments}

\begin{table*}[t]
\centering
\caption{
Separability evaluation results on single-skill tasks.
Each skill block reports the separability of the heads identified for that skill.
\textit{Avg.} is the average of the Patch- and Erasure-based separability scores, while \textit{Overall} averages the corresponding scores across the four skills.
Best and second-best results within each model are shown in \textbf{bold} and \underline{underline}, respectively.
}
\label{tab:single_skill_results}

\small
\setlength{\tabcolsep}{2pt}
\renewcommand{\arraystretch}{1.12}

\resizebox{\textwidth}{!}{%
\begin{tabular}{@{}ll*{15}{>{\centering\arraybackslash}m{0.95cm}}@{}}
\toprule

\multirow{2}{*}{\textbf{Model}}
& \multirow{2}{*}{\textbf{Method}}
& \multicolumn{3}{c}{\textbf{Counting Separability}}
& \multicolumn{3}{c}{\textbf{Position Separability}}
& \multicolumn{3}{c}{\textbf{Size Separability}}
& \multicolumn{3}{c}{\textbf{Distance Separability}}
& \multicolumn{3}{c}{\textbf{Overall}} \\

\cmidrule(lr){3-5}
\cmidrule(lr){6-8}
\cmidrule(lr){9-11}
\cmidrule(lr){12-14}
\cmidrule(lr){15-17}

& & Patch$\uparrow$ & Erasure$\uparrow$ & Avg.$\uparrow$
& Patch$\uparrow$ & Erasure$\uparrow$ & Avg.$\uparrow$
& Patch$\uparrow$ & Erasure$\uparrow$ & Avg.$\uparrow$
& Patch$\uparrow$ & Erasure$\uparrow$ & Avg.$\uparrow$
& Patch$\uparrow$ & Erasure$\uparrow$ & Avg.$\uparrow$ \\

\midrule

\multirow{4}{*}{\makecell[c]{Qwen2.5-VL\\7B-Instruct}}
& VFL \cite{shi2025vfl}
& -0.38 & -0.14 & -0.26
& -0.21 & -0.61 & -0.41
& -1.18 & -0.13 & -0.65
& 0.00 & -0.07 & -0.03
& -0.44 & -0.24 & -0.34 \\

& V-SEAM \cite{wang2025vseam}
& \underline{2.33} & -1.43 & \underline{0.45}
& \underline{23.17} & 1.26 & \underline{12.22}
& -36.85 & 0.27 & -18.29
& -16.38 & -2.33 & -9.35
& -6.93 & -0.56 & -3.74 \\

& NOTICE \cite{golovanevsky2025notice}
& -0.39 & \underline{0.19} & -0.10
& -0.21 & \underline{4.56} & 2.18
& \underline{4.36} & \underline{1.57} & \underline{2.97}
& \underline{15.64} & \underline{3.09} & \underline{9.37}
& \underline{4.85} & \underline{2.35} & \underline{3.60} \\

& Ours
& \textbf{55.03} & \textbf{28.87} & \textbf{41.95}
& \textbf{41.70} & \textbf{6.28} & \textbf{23.99}
& \textbf{26.82} & \textbf{13.29} & \textbf{20.05}
& \textbf{34.38} & \textbf{20.84} & \textbf{27.61}
& \textbf{39.48} & \textbf{17.32} & \textbf{28.40} \\

\midrule

\multirow{4}{*}{\makecell[c]{Qwen3-VL\\8B-Instruct}}
& VFL \cite{shi2025vfl}
& -0.54 & -2.50 & -1.52
& -0.38 & -0.71 & -0.54
& -0.95 & -0.08 & -0.51
& -1.22 & -1.80 & -1.51
& -0.77 & -1.27 & -1.02 \\

& V-SEAM \cite{wang2025vseam}
& 7.46 & \underline{17.06} & \underline{12.26}
& \underline{21.16} & \textbf{8.16} & \textbf{14.66}
& -28.64 & \underline{10.31} & -9.17
& 12.53 & 3.10 & 7.81
& 3.13 & 9.66 & 6.39 \\

& NOTICE \cite{golovanevsky2025notice}
& \underline{17.62} & 2.78 & 10.20
& -0.19 & \underline{-0.09} & -0.14
& \underline{6.01} & 1.76 & \underline{3.88}
& \underline{50.24} & \textbf{35.47} & \textbf{42.85}
& \underline{18.42} & \underline{9.98} & \underline{14.20} \\

& Ours
& \textbf{64.22} & \textbf{43.85} & \textbf{54.03}
& \textbf{21.68} & -4.09 & \underline{8.80}
& \textbf{26.27} & \textbf{23.56} & \textbf{24.91}
& \textbf{60.44} & \underline{24.72} & \underline{42.58}
& \textbf{43.15} & \textbf{22.01} & \textbf{32.58} \\

\midrule

\multirow{4}{*}{\makecell[c]{Gemma-4\\E4B-it}}
& VFL \cite{shi2025vfl}
& -0.98 & -1.69 & -1.33
& -0.08 & -10.85 & -5.46
& -1.85 & -1.34 & -1.60
& 0.21 & 1.29 & 0.75
& -0.67 & -3.15 & -1.91 \\

& V-SEAM \cite{wang2025vseam}
& 3.30 & -6.79 & -1.75
& 0.99 & \underline{1.32} & 1.15
& -11.60 & 1.04 & -5.28
& -30.28 & 8.68 & -10.80
& -9.40 & 1.06 & -4.17 \\

& NOTICE \cite{golovanevsky2025notice}
& \underline{8.21} & \underline{2.96} & \underline{5.58}
& \underline{2.99} & \textbf{8.01} & \underline{5.50}
& \underline{12.19} & \underline{1.07} & \underline{6.63}
& \underline{4.01} & \underline{10.79} & \underline{7.40}
& \underline{6.85} & \underline{5.71} & \underline{6.28} \\

& Ours
& \textbf{31.89} & \textbf{18.83} & \textbf{25.36}
& \textbf{23.18} & -3.81 & \textbf{9.69}
& \textbf{35.03} & \textbf{17.17} & \textbf{26.10}
& \textbf{24.85} & \textbf{13.11} & \textbf{18.98}
& \textbf{28.74} & \textbf{11.33} & \textbf{20.03} \\

\bottomrule
\end{tabular}%
}
\end{table*}












\subsection{Experimental Setup}

\vspace{0.1in}
\noindent \textbf{Models.}
We conduct comprehensive experiments on three instruction-tuned VLMs from different model families:
Qwen/Qwen2.5-VL-7B-Instruct \citep{bai2025qwen25vltechnicalreport},
Qwen/Qwen3-VL-8B-Instruct \citep{bai2025qwen3vltechnicalreport}, and
google/gemma-4-E4B-it \citep{gemmateam2026gemma4technicalreport}.

\vspace{0.1in}
\noindent \textbf{Data.}
We employ the SPHERE dataset~\citep{zhang2025sphere}, which identifies counting, positioning, size comparison, and distance comparison as fundamental skills for spatial understanding. SPHERE also demonstrates that combining these skills can solve a wide range of problems. The dataset covers both single-skill tasks and multi-skill tasks requiring two skills jointly. We split the single-skill data into two subsets: an identification set for head scoring and selection, and an evaluation set. Details on data statistics and examples are provided in the supplementary material.

\vspace{0.1in}
\noindent \textbf{Baselines.}
We compare with three head localization and skill localization methods for VLMs: V-SEAM~\citep{wang2025vseam}, NOTICE~\citep{golovanevsky2025notice}, and VFL~\citep{shi2025vfl}. For V-SEAM, we select the top-20 scoring heads per skill; for NOTICE, we select heads scoring more than two standard deviations above the skill mean and remove heads selected for multiple skills; for VFL, we select the highest-scoring layer per skill and intervene on all heads in that layer. 

\vspace{0.1in}
\noindent \textbf{Implementation details.}
We set the target-skill score threshold to $z_{\mathrm{th}}=0.3$ for Qwen models and $z_{\mathrm{th}}=0.1$ for the Gemma model. We consider separation threshold values $\Lambda=\{0, 0.5, 1.0, \ldots, 5.0\}$. Each random baseline is computed by averaging over $10$ trials. Random heads are sampled from the same layer range of the corresponding identified heads. For VFL \cite{shi2025vfl}, we select a random layer. For Erasure, we scale the head outputs by $0$, $-0.25$, and $-0.5$, and report the average effect.

\subsection{Main Results}
\label{sec:main_results}
\vspace{0.1in}
\noindent \textbf{Evaluating separability on single-skill tasks.}
We evaluate the separability of identified heads for each skill on the evaluation set of single-skill data, as defined in Eq.~(\ref{eq:separability}). Results are shown in Tab.~\ref{tab:single_skill_results}. The proposed method achieves the best overall separability on all three VLMs, with average scores of \textbf{28.40}, \textbf{32.58}, and \textbf{20.03} on Qwen2.5-VL, Qwen3-VL, and Gemma-4, respectively, exceeding the strongest baseline by \textbf{24.80}, \textbf{18.38}, and \textbf{13.75} points. Moreover, our method ranks first on \textbf{31} of the \textbf{36} skill-level metrics. These results show that our method, consisting of Surface-Form-Agnostic Head Scoring and Separable Head Selection, can effectively identify separable skill mechanisms in VLMs. We provide the detailed intermediate results and more analysis in the supplementary material.

Since the baselines identify heads using only natural images with the Content-based Option answer format, we additionally evaluate all methods on this single surface form. Our method still achieves the highest overall separability across all three VLMs, outperforming the strongest baselines by \textbf{29.09}, \textbf{17.99}, and \textbf{12.91} points, respectively. Detailed results are provided in the supplementary material.

\paragraph{Evaluating separability and composability on multi-skill tasks.}

\begin{table*}[t]
\centering
\caption{
Separability and composability evaluation results on multi-skill tasks.
\textit{Sep.} denotes the separability score, and \textit{Comp.} denotes the composability score. For each method, we determine its task-wise ranks for \textit{Sep.} and \textit{Comp.} among all methods within each model. We then compute the harmonic mean of these ranks separately for each metric, denoted as \textit{Rank}.
Best and second-best results within each model are shown in \textbf{bold} and \underline{underline}, respectively.
}
\label{tab:multiskill_results}

\small
\setlength{\tabcolsep}{2.5pt}
\renewcommand{\arraystretch}{1.12}

\begin{adjustbox}{max width=0.95\textwidth}
\begin{tabular}{@{}ll*{10}{>{\centering\arraybackslash}p{1.2cm}}@{}}
\toprule

\multirow{2}{*}{\textbf{Model}}
& \multirow{2}{*}{\textbf{Method}}
& \multicolumn{2}{c}{\makebox[0pt][c]{\textbf{Position + Size}}}
& \multicolumn{2}{c}{\makebox[0pt][c]{\textbf{Counting + Position}}}
& \multicolumn{2}{c}{\makebox[0pt][c]{\textbf{Distance + Size}}}
& \multicolumn{4}{c}{\textbf{Average}} \\

\cmidrule(lr){3-4}
\cmidrule(lr){5-6}
\cmidrule(lr){7-8}
\cmidrule(lr){9-12}

& &
\mbox{Sep.\,$\uparrow$} & \mbox{Comp.\,$\uparrow$}
& \mbox{Sep.\,$\uparrow$} & \mbox{Comp.\,$\uparrow$}
& \mbox{Sep.\,$\uparrow$} & \mbox{Comp.\,$\uparrow$}
& \mbox{Sep.\,$\uparrow$} & \mbox{Rank\,$\downarrow$}
& \mbox{Comp.\,$\uparrow$} & \mbox{Rank\,$\downarrow$} \\

\midrule

\multirow{4}{*}{\shortstack[c]{Qwen2.5-VL\\7B-Instruct}}
& VFL \cite{shi2025vfl}
& -0.52 & -0.87
& -3.09 & -4.70
& \underline{5.68} & -1.86
& 0.69 & 2.57
& -2.48 & 3.60 \\

& V-SEAM \cite{wang2025vseam}
& -3.11 & 0.00
& -4.08 & 0.00
& \textbf{7.35} & \underline{0.00}
& 0.05 & \underline{2.00}
& 0.00 & 2.57 \\

& NOTICE \cite{golovanevsky2025notice}
& \underline{0.26} & \underline{0.20}
& \underline{5.18} & \underline{1.11}
& 4.12 & \textbf{3.22}
& \underline{3.19} & 2.25
& \underline{1.51} & \underline{1.50} \\

& Ours
& \textbf{28.97} & \textbf{12.82}
& \textbf{29.26} & \textbf{8.28}
& -7.62 & -2.12
& \textbf{16.87} & \textbf{1.33}
& \textbf{6.33} & \textbf{1.33} \\

\midrule

\multirow{4}{*}{\shortstack[c]{Qwen3-VL\\8B-Instruct}}
& VFL \cite{shi2025vfl}
& 0.00 & \underline{0.00}
& 0.42 & -5.62
& -0.57 & -1.60
& -0.05 & 3.00
& -2.41 & 2.77 \\

& V-SEAM \cite{wang2025vseam}
& \underline{12.45} & -4.65
& -6.67 & -2.92
& \underline{4.84} & -2.32
& 3.54 & 2.40
& -3.30 & 3.60 \\

& NOTICE \cite{golovanevsky2025notice}
& -0.50 & \underline{0.00}
& \underline{4.38} & \underline{-0.10}
& \textbf{18.68} & \underline{7.85}
& \underline{7.52} & \underline{1.71}
& \underline{2.58} & \underline{2.00} \\

& Ours
& \textbf{32.70} & \textbf{13.65}
& \textbf{35.10} & \textbf{2.30}
& -2.52 & \textbf{8.47}
& \textbf{21.76} & \textbf{1.33}
& \textbf{8.14} & \textbf{1.00} \\

\midrule

\multirow{4}{*}{\shortstack[c]{Gemma-4\\E4B-it}}
& VFL \cite{shi2025vfl}
& \underline{12.20} & \underline{-0.14}
& \textbf{27.22} & \underline{-0.13}
& -11.73 & -10.85
& \textbf{9.23} & \underline{1.71}
& -3.71 & 2.40 \\

& V-SEAM \cite{wang2025vseam}
& -3.86 & -10.89
& \underline{25.41} & -1.80
& -0.65 & -6.46
& 6.97 & 2.77
& -6.38 & 3.60 \\

& NOTICE \cite{golovanevsky2025notice}
& 6.66 & -2.83
& -9.16 & \textbf{1.39}
& \textbf{27.78} & \underline{8.38}
& \underline{8.43} & 1.89
& \underline{2.31} & \underline{1.64} \\

& Ours
& \textbf{14.81} & \textbf{2.42}
& 5.37 & -1.53
& \underline{-0.31} & \textbf{8.92}
& 6.62 & \textbf{1.64}
& \textbf{3.27} & \textbf{1.29} \\

\bottomrule
\end{tabular}
\end{adjustbox}
\end{table*}

We evaluate identified heads on three multi-skill tasks: Position + Size, Counting + Position, and Distance + Size. The separability and composability metrics are defined in Eq.~(\ref{eq:multiskill_separability}) and Eq.~(\ref{eq:composability_rel}), respectively. The average score across three tasks can be dominated by an exceptionally high score on one task, even if performance on the other tasks is poor. To account for this, we also report the harmonic mean of the task-wise method ranks for each metric, which reflects whether the method consistently performs well on different tasks. Results are shown in Tab.~\ref{tab:multiskill_results}. 

Our method achieves the strongest overall  performance across all three VLMs. For composability, it achieves the highest average score and the best rank on all models, outperforming the strongest baselines by \textbf{4.82}, \textbf{5.56}, and \textbf{0.96} points on Qwen2.5-VL, Qwen3-VL, and Gemma-4, respectively. These results demonstrate that our method can effectively identify composable heads. For separability, our method achieves both the highest average score and the best rank on Qwen2.5-VL and Qwen3-VL, exceeding the strongest baselines by \textbf{13.68} and \textbf{14.24} points, respectively. Although its average separability on Gemma-4 is not the highest, our method achieves the best rank. This indicates that our method has consistently strong separability across the three tasks rather than merely performing exceptionally well on a single task. We show the detailed intermediate results in the supplementary material.

\subsection{Ablation Studies}

\vspace{0.1in}
\noindent \textbf{Effectiveness of separable head selection.}
We compare our method against three alternative head selection strategies on Qwen2.5-VL while keeping the head scoring procedure fixed. Top-$K$ selects the top-20 scoring heads for each skill, following prior works~\citep{bi2025unveiling,wang2025vseam}. Top-$K$ + Dedup. further removes heads selected for more than one skill. Global $\lambda^*$ uses the same head-selection criterion as ours but optimizes a single separation threshold shared across all skills by maximizing their average separability, yielding $\lambda=1.5$. Results are shown in Tab.~\ref{tab:adaptive_lambda_ablation}.

Top-$K$ performs worst, with negative separability on both single-skill and multi-skill evaluations. This indicates that selecting heads solely by their target-skill scores does not ensure separability. Top-$K$ + Dedup. substantially improves separability by removing shared heads. However, this strategy ignores the relative contributions of a head across different skills. It may discard heads that contribute much more to one skill than to other skills. Global $\lambda^*$ further improves separability by considering relative contributions across skills, but a shared separation threshold cannot adapt optimally to every skill. In contrast, our method optimizes a skill-specific $\lambda^*$ and achieves the highest separability across all metrics, reaching \textbf{28.40} on single-skill tasks and \textbf{16.87} on multi-skill tasks. These results demonstrate the importance of considering relative cross-skill contributions and adopting skill-specific separation thresholds for identifying separable heads.

\begin{table}[t]
\centering
\caption{
Evaluation results with different head selection strategies.
$\mathrm{Sep}_{\mathrm{Patch}}$ and $\mathrm{Sep}_{\mathrm{Eras}}$ denote the average separability across the four single-skill tasks, while multi-skill separability is averaged across the three multi-skill tasks.
Best and second-best results are shown in \textbf{bold} and \underline{underline}, respectively.
}
\label{tab:adaptive_lambda_ablation}

\footnotesize
\setlength{\tabcolsep}{4pt}
\renewcommand{\arraystretch}{1.12}

\resizebox{\columnwidth}{!}{%
\begin{tabular}{@{}lcccc@{}}
\toprule

\multirow{2}{*}{\textbf{Method}}
& \multicolumn{3}{c}{\textbf{Single-Skill}}
& \multicolumn{1}{c}{\textbf{Multi-Skill}} \\

\cmidrule(lr){2-4}
\cmidrule(lr){5-5}

&
$\mathrm{Sep}_{\mathrm{Patch}}$
& $\mathrm{Sep}_{\mathrm{Eras}}$
& Avg.
& Sep. \\

\midrule

Top-$K$ ($K=20$)
& -7.94
& -1.44
& -4.69
& -3.94 \\

Top-$K$ + Dedup.
& 13.04
& 3.76
& 8.40
& 0.26 \\

Global $\lambda^*$ ($\lambda=1.5$)
& \underline{30.08}
& \underline{16.99}
& \underline{23.54}
& \underline{16.20} \\

Skill-specific $\lambda^*$ (Ours)
& \textbf{39.48}
& \textbf{17.32}
& \textbf{28.40}
& \textbf{16.87} \\

\bottomrule
\end{tabular}%
}
\end{table}

\begin{table}[t]
\centering
\caption{
Separability evaluation results on single-skill tasks under different surface forms.
$\mathrm{Sep}_{\mathrm{Avg}}$ denotes the average of Overall Patch- and Erasure-based separability.
Best and second-best results are shown in \textbf{bold} and \underline{underline}, respectively.
}
\label{tab:surface_form_ablation}

\footnotesize
\setlength{\tabcolsep}{2.5pt}
\renewcommand{\arraystretch}{1.15}

\resizebox{\columnwidth}{!}{%
\begin{tabular}{
@{}
>{\raggedright\arraybackslash}m{3.65cm}
*{3}{>{\centering\arraybackslash}m{1.20cm}}
@{}
}
\toprule

& \multicolumn{3}{c}{
\textbf{Answer Format for Evaluation}
} \\

\cmidrule(lr){2-4}

\shortstack[l]{
\textbf{Answer Format}\\
\textbf{for Identification}
}
& \shortstack[c]{Content\\Options}
& \shortstack[c]{Label\\Options}
& \shortstack[c]{True\\or False} \\

\cmidrule(lr){2-4}

& $\mathrm{Sep}_{\mathrm{Avg}}$
& $\mathrm{Sep}_{\mathrm{Avg}}$
& $\mathrm{Sep}_{\mathrm{Avg}}$ \\

\midrule

Content Options
& \textbf{40.06}
& 6.67
& 8.10 \\

Label Options
& 11.10
& \textbf{18.49}
& \underline{8.55} \\

Content + Label (Ours)
& \underline{32.53}
& \underline{14.91}
& \textbf{10.78} \\

\midrule
\addlinespace[3pt]

& \multicolumn{3}{c}{
\textbf{Visual Style for Evaluation}
} \\

\cmidrule(lr){2-4}

\shortstack[l]{
\textbf{Visual Style}\\
\textbf{for Identification}
}
& Natural
& \shortstack[c]{Cartoon\\+ Oil}
& \shortstack[c]{Colored\\Sketch} \\

\cmidrule(lr){2-4}

& $\mathrm{Sep}_{\mathrm{Avg}}$
& $\mathrm{Sep}_{\mathrm{Avg}}$
& $\mathrm{Sep}_{\mathrm{Avg}}$ \\

\midrule

Natural
& \underline{28.48}
& 27.34
& 23.28 \\

Cartoon + Oil Painting
& 27.38
& \underline{28.66}
& \underline{24.36} \\

Natural + Cartoon + Oil (Ours)
& \textbf{29.36}
& \textbf{28.87}
& \textbf{24.53} \\

\bottomrule
\end{tabular}%
}
\end{table}

\vspace{0.1in}
\noindent \textbf{Effectiveness of surface-form-agnostic head scoring.}
To examine the robustness of our method to surface-form variations, we conduct ablation experiments on Qwen2.5-VL. We fix the visual style (answer format) while varying the answer format (visual style) used for head identification, and evaluate the identified heads across different surface forms. For OOD evaluation, we further introduce an unseen answer format, True or False, and an unseen visual style, Colored Sketch. In the True or False format, the model is given a question and a candidate answer and asked to judge its correctness. The evaluation results are shown in Tab.~\ref{tab:surface_form_ablation}.


For answer formats, single-format identification methods suffer substantial cross-format drops in separability. When evaluated on Content Options, Content-only identification achieves the highest separability of \textbf{40.06}, whereas Label-only identification achieves only \textbf{11.10} (\textbf{27.7\%} of the best score). Similarly, when evaluated on Label Options, Label-only identification achieves the highest separability of \textbf{18.49}, whereas Content-only identification achieves only \textbf{6.67} (\textbf{36.1\%} of the best score). These results suggest that using a single surface form can introduce spurious correlations between skills and surface forms.

In contrast, our method achieves \textbf{32.53} on Content Option and \textbf{14.91} on Label Option, reaching over \textbf{80\%} of the best separability under both formats. Moreover, on the unseen True or False format, our method achieves the highest separability. For visual styles, our method performs best in all evaluation settings. These results demonstrate the effectiveness of our surface-form-agnostic scoring.

\begin{table}[t]
\centering
\caption{
Evaluation results of different token-position strategies.
$\mathrm{Sep}_{\mathrm{Patch}}$ and $\mathrm{Sep}_{\mathrm{Eras}}$ denote the average separability across the four single-skill tasks.
For multi-skill tasks, Sep. and Comp. denote the average separability and composability across the three tasks, respectively. Best and second-best results are shown in \textbf{bold} and \underline{underline}, respectively.
}
\label{tab:token_position_ablation}

\footnotesize
\setlength{\tabcolsep}{2.5pt}
\renewcommand{\arraystretch}{1.12}

\begin{adjustbox}{max width=\columnwidth}
\begin{tabular}{@{}l*{5}{>{\centering\arraybackslash}p{1.15cm}}@{}}
\toprule

\multirow{2}{*}{\textbf{Token Positions}}
& \multicolumn{3}{c}{\textbf{Single-skill}}
& \multicolumn{2}{c}{\textbf{Multi-skill}} \\

\cmidrule(lr){2-4}
\cmidrule(lr){5-6}

& $\mathrm{Sep}_{\mathrm{Patch}}$
& $\mathrm{Sep}_{\mathrm{Eras}}$
& Avg.
& Sep.
& Comp. \\

\midrule

Final Position Only
& 24.88
& 8.85
& 16.86
& \underline{9.30}
& \underline{7.09} \\

Preceding Positions Only
& 33.52
& \underline{13.86}
& 23.69
& 9.25
& \textbf{7.46} \\

All Positions (Together)
& \underline{35.50}
& 12.47
& \underline{23.98}
& 3.66
& 6.52 \\

All Positions (Separate, Ours)
& \textbf{39.48}
& \textbf{17.32}
& \textbf{28.40}
& \textbf{16.87}
& 6.33 \\

\bottomrule
\end{tabular}
\end{adjustbox}
\end{table}

\vspace{0.1in}
\noindent \textbf{Effectiveness of token-position-aware head identification.}
\label{sec:token_position}
To examine the effectiveness of distinguishing preceding and final token positions during head identification, we compare our method with variants that use only the final position, only the preceding positions, or all positions together. We conduct experiments on Qwen2.5-VL, with results shown in Tab.~\ref{tab:token_position_ablation}.

Compared with our strategy, using only the final or preceding positions yields lower separability on both single-skill and multi-skill tasks. This suggests that skill-related heads are distributed across both position groups, and restricting localization to either one misses useful contributions from the other. Importantly, simply identifying heads using all token positions is not sufficient. Compared with our separate-position strategy, treating preceding and final positions together reduces single-skill separability from \textbf{28.40} to \textbf{23.98} and multi-skill separability from \textbf{16.87} to \textbf{3.66}. In contrast, composability remains relatively stable across all strategies, ranging from \textbf{6.33} to \textbf{7.46}. These results demonstrate that distinguishing head contributions across the two position groups substantially improves head separability while having a minor impact on composability.
\section{Conclusion}
\label{sec:conclusion}
In this paper, we propose a multi-skill interpretability framework, \longname{}. We posit that two desiderata, separability and composability, are critical for skill localization. Our framework not only identifies separable and composable heads but also introduces novel metrics to evaluate these properties. For head identification, we propose Separable Head Selection, which optimizes a separation threshold to select skill-specific heads based on their contributions across skills. To obtain reliable contribution scores, we introduce Surface-Form-Agnostic Head Scoring, which mitigates spurious correlations between skills and surface forms. Extensive experiments on single-skill and multi-skill tasks demonstrate that our method identifies more separable and composable heads than existing methods.
\section*{LLM Usage Statement}
We used GPT-5.4 to generate image-editing instructions and assist in checking answer consistency, and GPT-Image-1.5 to edit images and generate visual-style variants. Generative AI tools also assisted with method implementation, translation, and data cleaning and reformatting. The authors manually evaluated the quality and reliability of the generated data and verified AI-generated code. The authors take full responsibility for the final content of this work.

{
    \small
    \bibliographystyle{ieeenat_fullname}
    \bibliography{references}

@inproceedings{golovanevsky2025notice,
    title = "What Do {VLM}s {NOTICE}? A Mechanistic Interpretability Pipeline for {G}aussian-Noise-free Text-Image Corruption and Evaluation",
    author = "Golovanevsky, Michal  and
      Rudman, William  and
      Palit, Vedant  and
      Eickhoff, Carsten  and
      Singh, Ritambhara",
    editor = "Chiruzzo, Luis  and
      Ritter, Alan  and
      Wang, Lu",
    booktitle = "Proceedings of the 2025 Conference of the Nations of the Americas Chapter of the Association for Computational Linguistics: Human Language Technologies (Volume 1: Long Papers)",
    month = apr,
    year = "2025",
    address = "Albuquerque, New Mexico",
    publisher = "Association for Computational Linguistics",
    url = "https://aclanthology.org/2025.naacl-long.571/",
    doi = "10.18653/v1/2025.naacl-long.571",
    pages = "11462--11482",
    ISBN = "979-8-89176-189-6",
}

@inproceedings{wang2025vseam,
  title={V-seam: Visual semantic editing and attention modulating for causal interpretability of vision-language models},
  author={Wang, Qidong and Hu, Junjie and Jiang, Ming},
  booktitle={Proceedings of the 2025 Conference on Empirical Methods in Natural Language Processing},
  pages={17407--17431},
  year={2025}
}

@inproceedings{johnson2017clevr,
  title={Clevr: A diagnostic dataset for compositional language and elementary visual reasoning},
  author={Johnson, Justin and Hariharan, Bharath and Van Der Maaten, Laurens and Fei-Fei, Li and Lawrence Zitnick, C and Girshick, Ross},
  booktitle={Proceedings of the IEEE conference on computer vision and pattern recognition},
  pages={2901--2910},
  year={2017}
}

@inproceedings{hudson2019gqa,
  title={Gqa: A new dataset for real-world visual reasoning and compositional question answering},
  author={Hudson, Drew A and Manning, Christopher D},
  booktitle={2019 IEEE/CVF Conference on Computer Vision and Pattern Recognition (CVPR)},
  pages={6693--6702},
  year={2019},
  organization={IEEE}
}

@article{liu2023visual,
  title={Visual spatial reasoning},
  author={Liu, Fangyu and Emerson, Guy and Collier, Nigel},
  journal={Transactions of the Association for Computational Linguistics},
  volume={11},
  pages={635--651},
  year={2023},
  publisher={MIT Press One Broadway, 12th Floor, Cambridge, Massachusetts 02142, USA~…}
}

@article{zhao2022vl,
  title={Vl-checklist: Evaluating pre-trained vision-language models with objects, attributes and relations},
  author={Zhao, Tiancheng and Zhang, Tianqi and Zhu, Mingwei and Shen, Haozhan and Lee, Kyusong and Lu, Xiaopeng and Yin, Jianwei},
  journal={arXiv preprint arXiv:2207.00221},
  year={2022}
}

@article{yuksekgonul2022and,
  title={When and why vision-language models behave like bags-of-words, and what to do about it?},
  author={Yuksekgonul, Mert and Bianchi, Federico and Kalluri, Pratyusha and Jurafsky, Dan and Zou, James},
  journal={arXiv preprint arXiv:2210.01936},
  year={2022}
}

@inproceedings{zhang2025sphere,
  title={Sphere: Unveiling spatial blind spots in vision-language models through hierarchical evaluation},
  author={Zhang, Wenyu and Ng, Wei En and Ma, Lixin and Wang, Yuwen and Zhao, Junqi and Koenecke, Allison and Li, Boyang and Wang, Lu},
  booktitle={Proceedings of the 63rd Annual Meeting of the Association for Computational Linguistics (Volume 1: Long Papers)},
  pages={11591--11609},
  year={2025}
}

@inproceedings{xu2024multiskill,
  title={MultiSkill: Evaluating large multimodal models for fine-grained alignment skills},
  author={Xu, Zhenran and Shi, Senbao and Hu, Baotian and Wang, Longyue and Zhang, Min},
  booktitle={Findings of the Association for Computational Linguistics: EMNLP 2024},
  pages={1506--1523},
  year={2024}
}

@article{li2026unveiling,
  title={Unveiling the compositional ability gap in vision-language reasoning model},
  author={Li, Tianle and Zhang, Jihai and Rao, Yongming and Cheng, Yu},
  journal={Advances in Neural Information Processing Systems},
  volume={38},
  pages={126574--126592},
  year={2026}
}

@article{li2026two,
  title={Two Causally Related Needles in a Video Haystack},
  author={Li, Miaoyu and Chao, Qin and Li, Boyang},
  journal={Advances in Neural Information Processing Systems},
  volume={38},
  year={2025}
}

@article{kang2026characterizing,
  title={Characterizing Model-Native Skills},
  author={Kang, Feiyang and Dabas, Mahavir and Ko, Myeongseob and Jia, Ruoxi},
  journal={arXiv preprint arXiv:2604.17614},
  year={2026}
}

@inproceedings{tiong2024we,
  title={What are we measuring when we evaluate large vision-language models? an analysis of latent factors and biases},
  author={Tiong, Anthony and Zhao, Junqi and Li, Boyang and Li, Junnan and Hoi, Steven and Xiong, Caiming},
  booktitle={Proceedings of the 2024 Conference of the North American Chapter of the Association for Computational Linguistics: Human Language Technologies (Volume 1: Long Papers)},
  pages={3427--3454},
  year={2024}
}

@article{schaumloffel2026mechanisms,
  title={Mechanisms of Object Localization in Vision-Language Models},
  author={Schauml{\"o}ffel, Timothy and Vilas, Martina G and Roig, Gemma},
  journal={arXiv preprint arXiv:2605.19792},
  year={2026}
}

@article{che2026counting,
  title={Counting circuits: Mechanistic interpretability of visual reasoning in large vision-language models},
  author={Che, Liwei and Xue, Zhiyu and Quan, Yihao and Liu, Benlin and Shi, Zeru and Hurst, Michelle and Feldman, Jacob and Tang, Ruixiang and Krishna, Ranjay and Pavlovic, Vladimir},
  journal={arXiv preprint arXiv:2603.18523},
  year={2026}
}

@inproceedings{shi2025vfl,
 author = {Shi, Cheng and Yu, Yizhou and Yang, Sibei},
 booktitle = {Advances in Neural Information Processing Systems},
 doi = {10.52202/085713-0926},
 editor = {D. Belgrave and C. Zhang and H. Lin and R. Pascanu and P. Koniusz and M. Ghassemi and N. Chen},
 pages = {27510--27537},
 publisher = {Curran Associates, Inc.},
 title = {Vision Function Layer in Multimodal LLMs},
 url = {https://proceedings.neurips.cc/paper_files/paper/2025/file/27c4e15d9af120d7fef04432c7db577f-Paper-Conference.pdf},
 volume = {38, Main Conference},
 year = {2025}
}

@inproceedings{chen2026skill,
  title={Skill path: Unveiling language skills from circuit graphs},
  author={Chen, Hang and Yang, Xinyu and Zhu, Jiaying and Wang, Wenya},
  booktitle={Proceedings of the AAAI Conference on Artificial Intelligence},
  volume={40},
  pages={30210--30217},
  year={2026}
}

@article{jiang2025investigating,
  title={Investigating the functional roles of attention heads in vision language models: Evidence for reasoning modules},
  author={Jiang, Yanbei and Ma, Xueqi and Liu, Shu and Erfani, Sarah Monazam and Liu, Tongliang and Bailey, James and Lau, Jey Han and Ehinger, Krista A},
  journal={arXiv preprint arXiv:2512.10300},
  year={2025}
}

@article{zhu2025teach,
  title={How to teach large multimodal models new skills},
  author={Zhu, Zhen and Gong, Yiming and Xiao, Yao and Liu, Yaoyao and Hoiem, Derek},
  journal={arXiv preprint arXiv:2510.08564},
  year={2025}
}

@article{chen2023skill,
  title={Skill-it! a data-driven skills framework for understanding and training language models},
  author={Chen, Mayee and Roberts, Nicholas and Bhatia, Kush and Wang, Jue and Zhang, Ce and Sala, Frederic and R{\'e}, Christopher},
  journal={Advances in Neural Information Processing Systems},
  volume={36},
  pages={36000--36040},
  year={2023}
}

@inproceedings{zhao2025probing,
  title={Probing and boosting large language models capabilities via attention heads},
  author={Zhao, Dezhi and Liu, Xin and Feng, Xiaocheng and Wang, Hui and Qin, Bing},
  booktitle={Proceedings of the 2025 Conference on Empirical Methods in Natural Language Processing},
  pages={28518--28532},
  year={2025}
}

@inproceedings{bi2025unveiling,
  title={Unveiling visual perception in language models: An attention head analysis approach},
  author={Bi, Jing and Guo, Junjia and Tang, Yunlong and Wen, Lianggong Bruce and Liu, Zhang and Wang, Bingjie and Xu, Chenliang},
  booktitle={2025 IEEE/CVF Conference on Computer Vision and Pattern Recognition (CVPR)},
  pages={4135--4144},
  year={2025},
  organization={IEEE}
}

@misc{bai2025qwen25vltechnicalreport,
      title={Qwen2.5-VL Technical Report}, 
      author={Shuai Bai and Keqin Chen and Xuejing Liu and Jialin Wang and Wenbin Ge and Sibo Song and Kai Dang and Peng Wang and Shijie Wang and Jun Tang and Humen Zhong and Yuanzhi Zhu and Mingkun Yang and Zhaohai Li and Jianqiang Wan and Pengfei Wang and Wei Ding and Zheren Fu and Yiheng Xu and Jiabo Ye and Xi Zhang and Tianbao Xie and Zesen Cheng and Hang Zhang and Zhibo Yang and Haiyang Xu and Junyang Lin},
      year={2025},
      eprint={2502.13923},
      archivePrefix={arXiv},
      primaryClass={cs.CV},
      url={https://arxiv.org/abs/2502.13923}, 
}

@misc{bai2025qwen3vltechnicalreport,
      title={Qwen3-VL Technical Report}, 
      author={Shuai Bai and Yuxuan Cai and Ruizhe Chen and Keqin Chen and Xionghui Chen and Zesen Cheng and Lianghao Deng and Wei Ding and Chang Gao and Chunjiang Ge and Wenbin Ge and Zhifang Guo and Qidong Huang and Jie Huang and Fei Huang and Binyuan Hui and Shutong Jiang and Zhaohai Li and Mingsheng Li and Mei Li and Kaixin Li and Zicheng Lin and Junyang Lin and Xuejing Liu and Jiawei Liu and Chenglong Liu and Yang Liu and Dayiheng Liu and Shixuan Liu and Dunjie Lu and Ruilin Luo and Chenxu Lv and Rui Men and Lingchen Meng and Xuancheng Ren and Xingzhang Ren and Sibo Song and Yuchong Sun and Jun Tang and Jianhong Tu and Jianqiang Wan and Peng Wang and Pengfei Wang and Qiuyue Wang and Yuxuan Wang and Tianbao Xie and Yiheng Xu and Haiyang Xu and Jin Xu and Zhibo Yang and Mingkun Yang and Jianxin Yang and An Yang and Bowen Yu and Fei Zhang and Hang Zhang and Xi Zhang and Bo Zheng and Humen Zhong and Jingren Zhou and Fan Zhou and Jing Zhou and Yuanzhi Zhu and Ke Zhu},
      year={2025},
      eprint={2511.21631},
      archivePrefix={arXiv},
      primaryClass={cs.CV},
      url={https://arxiv.org/abs/2511.21631}, 
}

@misc{gemmateam2026gemma4technicalreport,
      title={Gemma 4 Technical Report}, 
      author={Gemma Team and Sherif El Abd and Vaibhav Aggarwal and Robin Algayres and Alek Andreev and Olivier Bachem and Ian Ballantyne and Cormac Brick and Victor Cărbune and Michelle Casbon and Mayank Chaturvedi and Aditya Chawla and Victor Cotruta and Alice Coucke and Phil Culliton and Robert Dadashi and Lucas Dixon and Mohamed Elhawaty and Utku Evci and Clément Farabet and Johan Ferret and Filippo Galgani and Sertan Girgin and Jean-Bastien Grill and Maarten Grootendorst and Jiaxian Guo and Cassidy Hardin and Yanzhang He and Steven M. Hernandez and Omri Homburger and Léonard Hussenot and Juyeong Ji and Armand Joulin and Aishwarya Kamath and Parnian Kassraie and Olivier Lacombe and Preethi Lahoti and Gaël Liu and Gus Martins and Luciano Martins and Tatiana Matejovicova and Ramona Merhej and Nikola Momchev and Sneha Mondal and Ryan Mullins and Sindhu Raghuram Panyam and Shreya Pathak and Sarah Perrin and André Susano Pinto and Etienne Pot and Angéline Pouget and Alexandre Ramé and Sabela Ramos and Douglas Reid and David Rim and Morgane Rivière and Karsten Roth and Louis Rouillard and Omar Sanseviero and Pier Giuseppe Sessa and Shane Settle and Danila Sinopalnikov and Sara Smoot and Piotr Stanczyk and Andreas Steiner and Lawrence Stewart and Ilya Tolstikhin and Michael Tschannen and Anton Tsitsulin and Nino Vieillard and Renjie Wu and Pingmei Xu and Haichuan Yang and Edouard Yvinec and Biao Zhang and Li Zhang and Joe Zou and Nicolas Aagnes and Abdelrahman Abdelhamed and Jakub Adamek and Shivani Agrawal and Shubham Agrawal and Ibrahim Alabdulmohsin and Jean Baptiste Alayrac and Uri Alon and Chandramouli Amarnath and Ankesh Anand and Chrysovalantis Anastasiou and Setareh Ariafar and François-Xavier Aubet and Kyriakos Axiotis and Federico Barbero and Joelle Barral and Alexei Bendebury and Urs Bergmann and Stanley Bileschi and Kat Black and Mathieu Blondel and Sebastian Borgeaud and Arthur Bražinskas and Ryan Burnell and Robert Busa-Fekete and Mu Cai and Daniele Calandriello and Glenn Cameron and Charlotte Caucheteux and Rahma Chaabouni and Garima Chadha and Jetha Chan and Blake Jianhang Chen and Jesse Chen and Lin Chen and Xu Chen and Derek Cheng and Tzu-hsiang Chien and Nikolai Chinaev and Yi Chou and Zhaohui Chu and Benjamin Coleman and Pooja Consul and Sam Conway-Rahman and Scott Crowell and Dylan Cutler and Vivek Dani and Samira Daruki and Anil Das and Daniel Deutsch and Nishanth Dikkala and Li Ding and Qiuhan Ding and Shenil Dodhia and Konstantin Donhauser and Tulsee Doshi and Anca Dragan and Alex Druinsky and Sahil Dua and Zoltan Egyed and Danielle Eisenbud and Daniel Eppens and Cindy Fan and Bahare Fatemi and Yassir Fathullah and Vlad Feinberg and Milen Ferev and Sebastian Flennerhag and Takumi Fujimoto and João Gabriel Oliveira and Isaac Galatzer-Levy and João Gante and Simon Geisler and Soham Ghosal and Antonious M. Girgis and Tamara von Glehn and Alec Go and Alhaad Gokhale and Alex Grills and Yiming Gu and Mayank Gupta and Pramod Gupta and Guru Guruganesh and Raia Hadsell and Hamza Harkous and Jitendra Harlalka and Demis Hassabis and Anja Hauth and Joe Heyward and Arian Hosseini and Chih-Yang Hsia and I-Hung Hsu and Xiaopeng Huang and Yangsibo Huang and Kevin Hui and Adrian Hutter and Te I and Fotis Iliopoulos and Advait Jain and Ganesh Jawahar and Ziwei Ji and Qilin Jin and Melvin Johnson and Kandarp Joshi and Arun Kandoor and Wang-Cheng Kang and Koray Kavukcuoglu and Mehran Kazemi and Kathleen Kenealy and Amr Khalifa and Phoebe Kirk and Ivan Korotkov and Suraj Kothawade and Vitaly Kovalev and Neel Kovelamudi and Adam Kraft and Ravin Kumar and Vivek Kumar and Harish Kuppam and Justin Lannin and Chen-Yu Lee and Seungji Lee and Dmitry Lepikhin and Alon Levkovitch and Dongdong Li and Qiujia Li and Valentin Liévin and Ethan Lin and Ziqian Lin and Casper Liu and Tianlin Liu and Tianqi Liu and Xin Liu and Ivan Lobov and Mayank Lunayach and Min Ma and Gagan Madan and Andrii Maksai and Eric Malmi and Michal Matuszak and Daniel McDuff and Gaurav Menghani and Maciej Mikuła and Daniil Mirylenka and Karolis Misiunas and Vedant Misra and Andreea Mitran and Kareem Mohamed and Maksim Mukha and Eric Noland and James O'Donnell and Brendan O'Donoghue and Kate Olszewska and Bernett Orlando and Wanqiong Pan and Rina Panigrahy and Unnati Parekh and Nicolas Perez-Nieves and Chunjong Park and Eric Paskie and Liqian Peng and Bryce Petrini and Slav Petrov and Jonas Pfeiffer and Bilal Piot and Martyna Plomecka and Siim Poder and Octavio Ponce and Arijit Pramanik and David Racz and Anish Rajan and Michelle Ramanovich and Anand Rao and Marvin Ritter and Vitor Rodrigues and Evan Rosen and Mikołaj Rybiński and Noveen Sachdeva and Michaël E. Sander and Rohit Sathyanarayana and Sagar Savla and Samuel Schmidgall and Tal Schuster and George Scrivener and Benoit Seguin and Andrew Sellergren and Aliaksei Severyn and Izhak Shafran and Dhruv Shah and Bobak Shahriari and Yuan Shangguan and Ashish Shenoy and Pradeep Shenoy and Rakesh Shivanna and Pauline Sho and Lucas Spangher and Wojciech Stokowiec and Tim Strother and Yao Su and Yinghao Sun and Mukund Sundararajan and Andrea Tacchetti and Mor Hazan Taege and Pouya Tafti and Jean Tarbouriech and Chetan Tekur and Shantanu Thakoor and Rahul Thapa and Madeleine Traverse and Lenart Treven and Tao Tu and Chien Te Tung and Çağlar Ünlü and Petar Veličković and Malini Pooni Venkat and Sagar Gubbi Venkatesh and Vidya Venkiteswaran and Francesco Visin and Alex Vitvitskyi and Kiran Vodrahalli and Weiyi Wang and Xin Wang and Tris Warkentin and Jan Wassenberg and John Wieting and Cindy Wu and Lechao Xiao and Hao Xu and Yuhui Xu and Fuzhao Xue and Arun Yadav and Jun Yan and Antoine Yang and Lin Yang and Ming-Hsuan Yang and Ziyu Ying and Jae Hyeon Yoo and Morteza Zadimoghaddam and Sajjad Zafar and Fred Zhang and Jiageng Zhang and Jianyi Zhang and Xiaofan Zhang and Chao Zhao and David Zhou and Chen Zou},
      year={2026},
      eprint={2607.02770},
      archivePrefix={arXiv},
      primaryClass={cs.CL},
      url={https://arxiv.org/abs/2607.02770}, 
}

@misc{ismithdeen2025promptception,
      title={Promptception: How Sensitive Are Large Multimodal Models to Prompts?}, 
      author={Mohamed Insaf Ismithdeen and Muhammad Uzair Khattak and Salman Khan},
      year={2025},
      eprint={2509.03986},
      archivePrefix={arXiv},
      primaryClass={cs.CV},
      url={https://arxiv.org/abs/2509.03986}, 
}

@inproceedings{atabuzzaman2025benchmarking,
  title={Benchmarking and mitigating mcqa selection bias of large vision-language models},
  author={Atabuzzaman, Md and Asgarov, Ali and Thomas, Chris},
  booktitle={Proceedings of the 2025 Conference on Empirical Methods in Natural Language Processing},
  pages={33536--33550},
  year={2025}
}

@article{rosenthal2025unexplored,
  title={Unexplored flaws in multiple-choice VQA evaluations},
  author={Rosenthal, Fabio and Schmidt, Sebastian and Graf, Thorsten and Bagodonat, Thorsten and G{\"u}nnemann, Stephan and Schwinn, Leo},
  journal={arXiv preprint arXiv:2511.22341},
  year={2025}
}

@article{vig2020causal,
  title={Causal mediation analysis for interpreting neural nlp: The case of gender bias},
  author={Vig, Jesse and Gehrmann, Sebastian and Belinkov, Yonatan and Qian, Sharon and Nevo, Daniel and Sakenis, Simas and Huang, Jason and Singer, Yaron and Shieber, Stuart},
  journal={arXiv preprint arXiv:2004.12265},
  year={2020}
}

@article{meng2022locating,
  title={Locating and editing factual associations in gpt},
  author={Meng, Kevin and Bau, David and Andonian, Alex and Belinkov, Yonatan},
  journal={Advances in neural information processing systems},
  volume={35},
  pages={17359--17372},
  year={2022}
}

@article{wang2022interpretability,
  title={Interpretability in the wild: a circuit for indirect object identification in GPT-2 small},
  author={Wang, Kevin and Variengien, Alexandre and Conmy, Arthur and Shlegeris, Buck and Steinhardt, Jacob},
  journal={arXiv preprint arXiv:2211.00593},
  year={2022}
}

@article{conmy2023towards,
  title={Towards automated circuit discovery for mechanistic interpretability},
  author={Conmy, Arthur and Mavor-Parker, Augustine and Lynch, Aengus and Heimersheim, Stefan and Garriga-Alonso, Adri{\`a}},
  journal={Advances in Neural Information Processing Systems},
  volume={36},
  pages={16318--16352},
  year={2023}
}

@inproceedings{zhang2024towards,
  title={Towards best practices of activation patching in language models: Metrics and methods},
  author={Zhang, Fred and Nanda, Neel},
  booktitle={International Conference on Learning Representations},
  volume={2024},
  pages={1651--1678},
  year={2024}
}

@article{heimersheim2024use,
  title={How to use and interpret activation patching},
  author={Heimersheim, Stefan and Nanda, Neel},
  journal={arXiv preprint arXiv:2404.15255},
  year={2024}
}

@inproceedings{comosition2,
  title={Circuit compositions: Exploring modular structures in transformer-based language models},
  author={Mondorf, Philipp and Wold, Sondre and Plank, Barbara},
  booktitle={Proceedings of the 63rd Annual Meeting of the Association for Computational Linguistics (Volume 1: Long Papers)},
  pages={14934--14955},
  year={2025}
}

@article{composition3,
  title={From Reasoning Traces to Reusable Modules: Understanding Compositional Generalization in Language Model Reasoning},
  author={Kong, Lingjing and Liu, Xin and Chen, Guangyi and Ma, Martin Q and Song, Xiangchen and Sun, Yuekai and Yurochkin, Mikhail and Killian, Taylor W and Salakhutdinov, Ruslan and Zhang, Kun and others},
  journal={arXiv preprint arXiv:2606.18089},
  year={2026}
}
}

\end{document}